\documentclass{article}

\usepackage{microtype}
\usepackage{graphicx}
\usepackage{subfigure}
\usepackage{booktabs} 
\usepackage{subfiles}
\usepackage{flafter}

\usepackage{hyperref}

\usepackage{soul}
\usepackage{xcolor}
\usepackage{graphicx}
\definecolor{electriclavender}{rgb}{0.91, 0.73, 1.0}
\definecolor{lightcyan}{rgb}{0.88, 1.0, 1.0}
\definecolor{dawnblue}{rgb}{0.84, 0.92, 1.0}

\newcommand{\best}[1]{\colorbox{dawnblue}{\bf #1}}
\newcommand{\otrbest}[1]{\colorbox{lightcyan}{\bf #1}}
\definecolor{azure(colorwheel)}{rgb}{0.0, 0.5, 1.0}
\definecolor{azurelight}{rgb}{0.0, 0.5, 0.8}

\usepackage{color, colortbl}
\definecolor{Gray}{gray}{0.9}
\usepackage{hyperref}
 \hypersetup{
     colorlinks=true,
     linkcolor=azurelight,
     filecolor=azurelight,
     citecolor = azurelight,
     urlcolor=azurelight,
 }

\usepackage[accepted]{icml2021}

\icmltitlerunning{Exploring Corruption Robustness: Inductive Biases in Vision Transformers and MLP-Mixers}

\begin{document}

\twocolumn[
\icmltitle{Exploring Corruption Robustness: Inductive Biases in Vision Transformers and MLP-Mixers}
\icmlsetsymbol{equal}{*}
\begin{icmlauthorlist}
\icmlauthor{Katelyn Morrison}{equal,to}
\icmlauthor{Benjamin Gilby}{equal,to}
\icmlauthor{Colton Lipchak}{to}
\icmlauthor{Adam Mattioli}{to}
\icmlauthor{Adriana Kovashka}{to}
\end{icmlauthorlist}
\icmlaffiliation{to}{Department of Computer Science, University of Pittsburgh, Pittsburgh, United States}
\icmlcorrespondingauthor{Katelyn Morrison}{kmorrison@pitt.edu}
\icmlcorrespondingauthor{Benjamin Gilby}{beg59@pitt.edu}
\icmlcorrespondingauthor{Colton Lipchak}{cjl99@pitt.edu}
\icmlcorrespondingauthor{Adam Mattioli}{afm45@pitt.edu}
\icmlcorrespondingauthor{Adriana Kovashka}{kovashka@cs.pitt.edu}
\icmlkeywords{Computer Vision, Robustness, Transformers}

\vskip 0.3in]



\printAffiliationsAndNotice{\icmlEqualContribution} 

\begin{abstract}
Recently, vision transformers and MLP-based models have been developed in order to address some of the prevalent weaknesses in convolutional neural networks. Due to the novelty of transformers being used in this domain along with the self-attention mechanism, it remains unclear to what degree these architectures are robust to corruptions. Despite some works proposing that data augmentation remains essential for a model to be robust against corruptions, we propose to explore the impact that the architecture has on corruption robustness. We find that vision transformer architectures are inherently more robust to corruptions than the ResNet-50 and MLP-Mixers. 
We also find that vision transformers with 5 times fewer parameters than a ResNet-50 have more shape bias. Our
\href{https://github.com/katelyn98/CorruptionRobustness}{code} is available to reproduce.
\end{abstract}

\section{Introduction}

 Research indicates that humans tend to classify objects based on shape rather than color or texture while convolutional neural networks are more biased towards texture \citep{ritter2017cognitive}. Developing and deploying reliable, accurate computer vision models is integral to the success and trust of vision-based technologies such as self-driving cars or assistive technologies.
 Significant errors in these technologies could be fatal, which is why it is important to understand the limitations of different models. 
 
 In the past decade, convolutional neural networks (CNNs) have been the state-of-the-art for computer vision tasks such as image classification. 
However, recent research has shown the limitations of CNNs in domain generalization tasks. Several recent works seeking to develop and train models that can successfully achieve domain adaptation and/or domain generalization investigate the intricacies that contribute to corruption robustness in a model \citep{mummadi2021does, DBLP:journals/corr/abs-1911-09071, geirhos2018imagenettrained, DBLP:journals/corr/abs-1907-12892, DBLP:journals/corr/abs-1802-02745}. Recently, novel architectures have achieved exceptional performance on ImageNet and CIFAR baselines. While robustness of CNNs have been studied, it is vital to explore the robustness of these new architectures, including how well they perform when presented corrupted images. 
 
 \paragraph{Contributions.} In this work, 
we investigate models with three different types of architectures: CNNs, Vision Transformers, and MLP-Mixers. In total, we compare and contrast twenty different pre-trained models.
 Our findings reveal how various pre-trained vision transformer architectures and MLP-Mixers make decisions (i.e., based on shape or texture) and how well they handle corruptions. Our contributions include the following:

\begin{itemize}
    \item Comparing corruption robustness and shape bias across state-of-the-art vision transformer architectures and the MLP-Mixer.
    \item Showing that vision transformers are inherently more robust to common corruptions than CNNs and the MLP-Mixer.
\end{itemize}

\section{Related Works}

Investigating inductive biases, such as shape bias and texture bias, and how these biases can improve the robustness of a model have been extensively explored within CNNs. We will highlight several advancements and contributions in the past few years ranging from data augmentation techniques to novel architectures designed to improve top-1 and top-5 accuracy on image classification tasks. 

\subsection{Data Augmentation \& Training Techniques}

\citet{geirhos2018imagenettrained} conduct an empirical study to understand the inductive biases learned by CNNs trained on ImageNet. After creating several augmented ImageNet data sets, they show that CNNs are more texture-biased during object recognition tasks while humans are more shape biased. These results are contradicted soon after showing that CNNs can learn shape bias as easily as texture bias \citep{DBLP:journals/corr/abs-1911-09071}. \citet{DBLP:journals/corr/abs-1911-09071} indicate that the inductive biases that the CNN learns may be solely dependent on the data it sees instead of the architecture itself. 

A more recent empirical study investigates if shape bias and corruption robustness have a direct correlation \citep{mummadi2021does}. \citet{mummadi2021does} compares the accuracy and corruption robustness of CNNs trained on ImageNet with standard images, standard and stylized images, and a combination of edge maps of ImageNet and standard images. They show that the model trained on standard images and edge maps resulted in having the greatest shape bias. However, the network trained on standard and stylized images performed the best on common corruptions. They concluded that the stylized images caused increased shape bias, but corruption robustness was increased by the stylized images, not the shape bias directly. 

An alternative approach explains an algorithm for shape-texture debiased learning by augmenting images in the training set with conflicting shapes and textures \citep{li2021shapetexture}. This algorithm is still based on using CNNs, but their algorithm proves to achieve improvements on ImageNet-C and Stylized-ImageNet among others. 
The augmentation in this algorithm consists of using conflicting shape and texture information on the original image.

\subsection{Architectures for Better Accuracy}

Different convolutional neural network architectures have been modified and reconstructed to achieve a higher accuracy on image classification tasks. Most recently, transformers have been modified and adapted for vision tasks such as image classification. We will only introduce the vision transformer architectures that we included in our experiments, but there are several other variations of vision transformers in the literature.


An architecture called the Vision Transformer (ViT) uses layers of multi-headed self attention and multi-layer perceptrons \citep{dosovitskiy2021an}. They conduct image classification by splitting an image into a fixed number of patches and embedding each image patch. This architecture achieves excellent results compared to CNNs on numerous baselines.  \citet{bhojanapalli2021understanding} investigate several different ViT and ResNet models to understand the robustness of the ViT models. They also show how the two architectures perform when faced against different adversarial attacks such as PGD and FGSM. Overall, their results reveal that the ViT is as least as robust to corruptions as ResNets \citep{bhojanapalli2021understanding}.

A variation of the ViT vision transformer, called the Swin Transformer, calculates self-attention of a window of image patches to compute predictions for tasks such as image classification \citep{liu2021swin}. The windows of image patches shift after calculating the self-attention of the previous windows. This shift results in a hierarchical feature map that provides a better global representation of the image. 

Two other variations of the vision transformer architectures are the Data-efficient Image Transformers (DeiT) \citep{touvron2020deit} and Class-Attention in Image Transformers (CaiT) \citep{touvron2021going}. DeiT uses a custom distillation procedure and no convolutional layers and CaiT features class-attention layers. 

A recent architecture, called MLP-Mixer, is designed to exclude convolutional and self-attention layers, and instead mix per-location features and spatial information through two MLP-based layers \citep{tolstikhin2021mlp}. This architecture also incorporates significant augmentation within the pre-processing pipeline to increase accuracy of the model. 

To our knowledge, no current research has been published on how numerous different vision transformers compare to one another in terms of corruption robustness or shape bias. No previous research has explored how robust the MLP-Mixer is to corruption either. 

\section{Method}

To explore how robust vision transformer architectures and MLP-Mixers are to corruptions, we perform several experiments consisting of four pre-trained CNNs, fourteen vision transformers, and two MLP-Mixers.

\subsection{Pre-trained Models}

\paragraph{Convolutional Neural Networks.}
The convolutional neural networks we chose were inspired by the models used in \citet{geirhos2018imagenettrained}. Specifically, \citet{geirhos2018imagenettrained} evaluated the shape bias of ResNet-50 \citep{he2015deep}, AlexNet \citep{NIPS2012_c399862d}, VGG-16 \citep{simonyan2015deep}, and GoogLeNet \citep{szegedy2014going}.  We evaluated the  and corruption robustness of these models to use as a baseline when determining how the vision transformers and the MLP-Mixers perform. 
 
 \paragraph{MLP-Mixers.} We evaluated two different variations of the MLP-Mixer architecture: the base and large variations. These pre-trained models were provided by the \texttt{timm library} \citep{rw2019timm}. 

\paragraph{Vision Transformers.}
We evaluated a total of four state-of-the-art, competing vision transformers. Due to limited resources and ease of access, we choose to use the pre-trained models provided by the \texttt{timm library} \citep{rw2019timm}. 
Specifically from \citet{rw2019timm}, we included the Swin-T, the ViT, and the CaiT pre-trained models in our evaluation. Each of these architectures have multiple pre-trained models available. We used four different pre-trained Swin Transformers, two different pre-trained ViT models, and two different pre-trained CaiT models. We obtained pre-trained DeiT models directly from \href{https://github.com/facebookresearch/deit?fbclid=IwAR2qzERDHwVdKSlah1v7MCsqp15EigeAjZbYp1F_YHm3ZR2-Bxkcejmq5r0}{Facebook Research's GitHub} \cite{touvron2020deit}. We used six different pre-trained DeiT models.



\subsection{Data Sets}

\paragraph{ImageNet-C.} We evaluated all of our pre-trained models on ImageNet-C to determine corruption robustness \citep{DBLP:journals/corr/abs-1807-01697}. ImageNet-C is a benchmark dataset used to assess how robust a model is to common corruptions. This dataset consists of nineteen different corruption types that are categorized within five general corruption categories (\textit{blur, weather, noise, digital, and extra}) for five different severity levels. The dataset is built off of the ILSVRC 2012 validation set which has $1,000$ classes 
and fifty validation images for each class
totalling $50,000$ validation images. In terms of ImageNet-C, each corruption type (i.e., blur $\rightarrow$ motion\_blur) has $50,000$ images for each severity level.

\paragraph{Texture-Cue Conflict.} We used the Texture-Cue Conflict dataset from \citet{geirhos2018imagenettrained} to evaluate the shape bias of our models. The Texture-Cue Conflict dataset consists of images that have the shape of one class combined with the texture of another. This results in conflicting shape and texture in each image. Two labels are included to identify ground-truth for \textit{both} the shape and the texture of an image. The dataset includes 16 classes and 80 images per class for a total of 1280 images.

\subsection{Evaluation Metrics}
The top-1 accuracy and top-1 error from each model is used to understand how robust the model is to different corruptions. Since we are evaluating architectures that are significantly different than CNNs, we decided to deviate away from using the corruption error from AlexNet as a normalization factor \cite{hendrycks2018benchmarking} when calculating the corruption error. Instead, we obtain the corruption error, $CE$, by sum the top-1 error for that corruption from severity 2 and severity 5 where $f$ is the given model, $s$ is the severity, and $c$ is the corruption:

\begin{center}
$
CE^{f}_{c} = E^{f}_{2,c} - E^{f}_{5,c}
$
\end{center}

To calculate mean corruption error, \textbf{mCE}, we take the average over all corruption errors calculated for a given model. Typically, mCE is calculated by averaging over corruption errors from all severity levels, but we chose to only include the corruption errors from severity 2 and severity 5 in our mCE calculation. We use these two severity levels to represent an average of the overall mCE for a given model. Even though our resulting mCE will not be directly comparable to previously published mCEs, it still provides enough evidence to draw conclusions about the models we evaluated.

We provide the top-1 accuracy on ILSVRC 2012 validation images \citep{ILSVRC15} because this is the dataset used for creating ImageNet-C. This metric will help us understand how the model performs on a dataset without corruptions.

Each pre-trained model was also evaluated on the texture-cue conflict dataset from \citet{geirhos2018imagenettrained} to calculate shape bias. The shape bias of a model is how much the model depends on shape when classifying images while texture bias is how much the model depends on the texture. shape bias, as stated by \citet{geirhos2018imagenettrained}, is calculated by the following formula:

\begin{center}
$ SB = shape_{correct} / (shape_{correct} + texture_{correct})$
\end{center}

\section{Results}

By evaluating twenty different pre-trained models on a subset of ImageNet-C and on the Texture-Cue Conflict dataset, we expose the robustness and inductive biases for each of these models. Please refer to our appendix for more in depth results from our experiments.

\subsection{Corruption Robustness}

We evaluated every pre-trained model on ImageNet-C and calculated the mCE to understand how each model performed against common corruptions in Table \ref{corruption}. 
We group the pre-trained models by the type of architecture. For example, the ResNet-50, AlexNet, GoogLeNet, and VGG16 are all types of CNNs and are grouped together in Table \ref{corruption}. 
When referring to Table \ref{corruption}, a lower mCE is more favorable and a higher top-1 accuracy is more favorable.

\begin{table}[t]
\caption{Evaluating Convolutional Neural Networks against Vision Transformer Architectures and MLP-Mixers on ImageNet-C. mCE is calculated using only severity 2 and 5. Top-1 accuracy is calculated for ILSVRC 2012 validation set.} 
\label{corruption}
\vskip 0.15in
\begin{center}
\begin{small}
\begin{sc}
\begin{tabular}{lcccr}
\toprule
 \multicolumn{3}{c}{Convolutional Neural Networks} \\
\toprule
Model &  top-1(\%) & mCE(\%)  &  \#params(M)\\
\midrule
ResNet-50 & $76.02$ & $65.54$ & $26$ \\
AlexNet  & $56.44$ & $83.18$ & $61$ \\
GoogLeNet  & $71.70$ & $68.82$ & $7$ \\
VGG-16  & $69.63$ & $75.10$ & $138$ \\
\bottomrule
\end{tabular}

\medskip

\begin{tabular}{lcccr}
\toprule
 \multicolumn{3}{c}{MLP-Mixers} \\
\toprule
Model & top-1(\%) & mCE(\%)  &  \#params(M)\\
\midrule
MLP-Mixer\_B   & $72.53$ & $65.54$ & $60$ \\
MLP-Mixer\_L  & $68.25$ & $69.65$ & $208$ \\
\bottomrule
\end{tabular}
\medskip

\begin{tabular}{lccr}
\toprule
\multicolumn{3}{c}{Vision Transformer Architectures} \\
\toprule
Model & top-1(\%) & mCE(\%)  &  \#params(M)\\
\midrule
ViT\_base  & $75.73$ & $58.55$ & $86$\\
ViT\_large  & $79.16$ & $49.02$ & $304$ \\
DeiT\_base  &  $81.84$ & $42.30$ & $86$ \\
DeiT\_base-dist.  & $83.16$ & $41.19$ & $87$\\
DeiT\_small & $79.68$ & $47.79$ & $22$ \\
DeiT\_small-dist. & $81.05$ & $46.25$ & $22$ \\
DeiT\_tiny  & $71.92$ & \otrbest{60.08} & \otrbest{5}\\
DeiT\_tiny-dist. & $74.38$ & $57.45$ & $6$ \\
CaiT\_s24  & $83.28$ & $40.59$ & $47$ \\
CaiT\_xxs24  & $78.38$ & $49.28$ & $11$ \\
Swin-T\_tiny  & $80.85$ & $50.70$ & $28$ \\
Swin-T\_small & $82.96$ & $45.51$ & $50$ \\
Swin-T\_base  & $84.90$ & $38.52$ & $88$ \\
Swin-T\_large  & $85.92$ & \best{34.63} & \best{197} \\

\bottomrule

\end{tabular}
\end{sc}
\end{small}
\end{center}
\vskip -0.1in
\end{table}

We observe that MLP-Mixer models perform similarly to the CNNs when tested on ImageNet-C. All of the vision transformer models we evaluated achieved a significantly better mCE than the MLP-Mixers and CNNs. One significant observation is that the tiny DeiT vision transformer with only five million parameters achieves an mCE of $60.08$\% while a ResNet50 with approximately five times the parameters has an mCE of $65.54$\%. 
Overall, the model that achieved the lowest mCE at $34.63$\% was the large Swin Transformer with $197$ million parameters. This model also performed the best on the uncorrupted ILSVRC 2012 validation set with a top-1 accuracy of $85.92$\%. We suspect the Swin transformer performed the best out of all of the vision transformers because of its shifting windows feature providing a global representation of the image.  

\subsection{Shape Bias}

We evaluated every pre-trained model on the Texture-Cue Conflict dataset and calculated shape bias to understand whether models were biased towards shape or texture when making decisions.
When referring to Table \ref{sample-table}, a higher shape bias is more favorable.
 
\begin{table}[t]
\caption{Evaluating shape bias of Convolutional Neural Networks against Vision Transformer Architectures and MLP-Mixers on Texture-Cue Conflict dataset.}
\label{sample-table}
\vskip 0.15in
\begin{center}
\begin{small}
\begin{sc}
\begin{tabular}{lcccr}
\toprule
 \multicolumn{3}{c}{Convolutional Neural Networks} \\
\toprule
Model & Shape Bias (\%) & \# params (M) \\
\midrule
ResNet-50    & $26.17$ & $26$ \\
AlexNet  & $29.80$ & $61$ \\
GoogLeNet & $28.52$ & $7$ \\
VGG-16 & $16.12$ & $138$ \\
\bottomrule
\end{tabular}

\medskip

\begin{tabular}{lcccr}
\toprule
 \multicolumn{3}{c}{MLP-Mixers} \\
\toprule
Model & Shape Bias (\%) & \# params (M) \\
\midrule
MLP-Mixer\_base  & $36.90$ & $60$ \\
MLP-Mixer\_large  & $38.64$ & $208$ \\
\bottomrule
\end{tabular}
\medskip

\begin{tabular}{lccr}
\toprule
\multicolumn{3}{c}{Vision Transformer Architectures} \\
\toprule
Model & Shape Bias (\%) & \# params (M) \\
\midrule
ViT\_base & $49.10$ & $86$ \\
ViT\_large & \best{55.35} & \best{304}\\
DeiT\_base & $42.32$ & $86$\\
DeiT\_base-dist. & $39.62$ & $87$\\
DeiT\_small & $38.26$ & $22$\\
DeiT\_small-dist. & $36.65$ & $22$\\
DeiT\_tiny & \otrbest{29.37} & \otrbest{5}\\
DeiT\_tiny-dist. & $31.06$ & $6$\\
CaiT\_s24 & $38.65$ & $47$\\
CaiT\_xxs24 & $34.24$ & $11$\\
Swin-T\_tiny & $25.21$ & $28$ \\
Swin-T\_small & $27.43$ & $50$ \\
Swin-T\_base & $36.39$ & $88$ \\
Swin-T\_large & $40.20$ & $197$ \\

\bottomrule
\end{tabular}
\end{sc}
\end{small}
\end{center}
\vskip -0.1in
\end{table}

We observe that the MLP-Mixers and vision transformers are more biased towards shape than CNNs, and many of the vision transformer models perform similarly to the MLP-Mixers. 
Notably, the tiny Data-efficient image transformer (DeiT\_tiny) architecture with approximately five times fewer parameters than a ResNet-50 achieves a shape bias of $29.37$\% compared to $26.17$\% for the ResNet50. The best performing vision transformer was the large ViT model with a shape bias of $55.35$\% and $304$ million parameters. 


Table \ref{corruption} and Table \ref{sample-table} highlight a general inverse relationship between shape bias and mean corruption error. As a model is more robust to common corruptions (smaller mCE), its shape bias increases.
We do not observe any relationship between the shape bias or mCE and the number of parameters. 

\section{Conclusion and Future Work}
 
 We compare several state-of-the-art vision transformers against CNNs and MLP-Mixers to better understand how these different architectures handle corruptions and if they rely on shape or texture more when classifying images. As seen in the graph in our appendix, we generally observe that when a model has a strong bias towards shape, it is more robust to common corruptions such as those in ImageNet-C. This conclusion concurs with \citet{geirhos2018imagenettrained}. 

Future directions include incorporating the rest of the severity levels from ImageNet-C to calculate the final mean corruption error for each model. It would also be beneficial to investigate different datasets such as ImageNet-A \citep{hendrycks2021natural}, ImageNet-P \citep{hendrycks2018benchmarking}, and ImageNet-R \citep{hendrycks2020many} to narrow down which specific components of these architectures are robust against all corruptions and perturbations.

\newpage
\bibliography{references}
\bibliographystyle{icml2021}

\pagebreak


\appendix


\clearpage

\begin{figure*}[htp!]
    \centering
    \includegraphics[width=12cm, height=6cm]{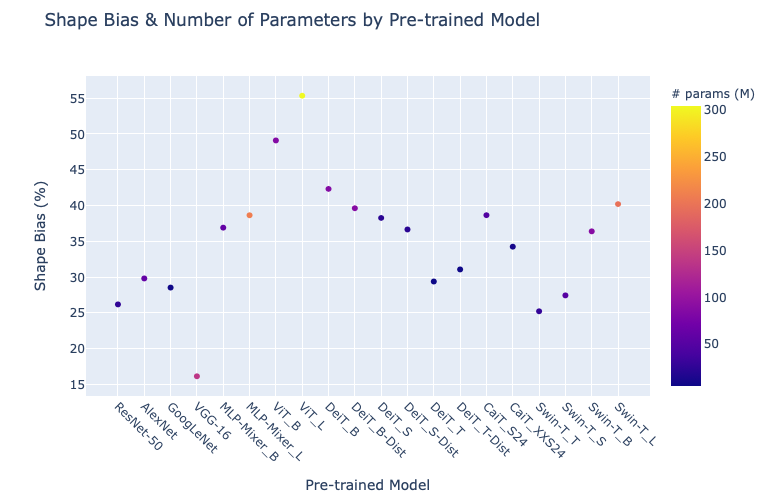}
    \caption{Shape Bias \& Number of Parameters by Pre-trained Model}
    \label{fig:graph1}
\end{figure*}

\begin{figure*}[htp!]
    \centering
    \includegraphics[width=12cm,height=6cm]{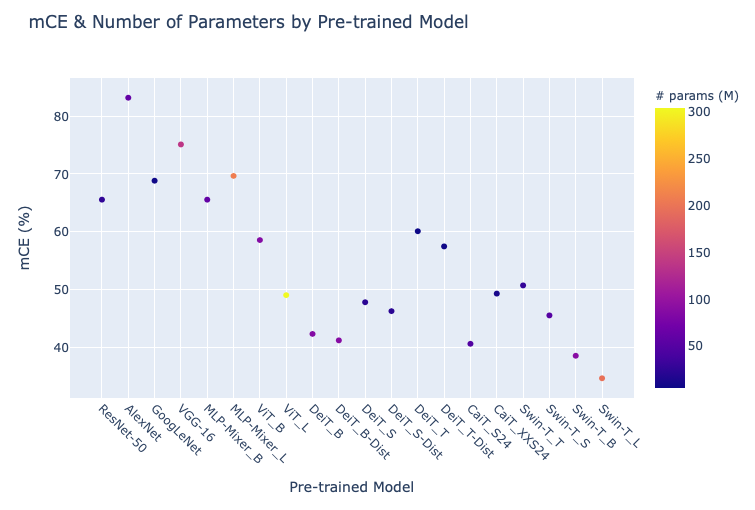}
    \caption{Mean Corruption Error \& Number of Parameters by Pre-trained Model}
    \label{fig:graph2}
\end{figure*}

\begin{figure*}[htp!]
    \centering
    \includegraphics[width=12cm,height=6cm]{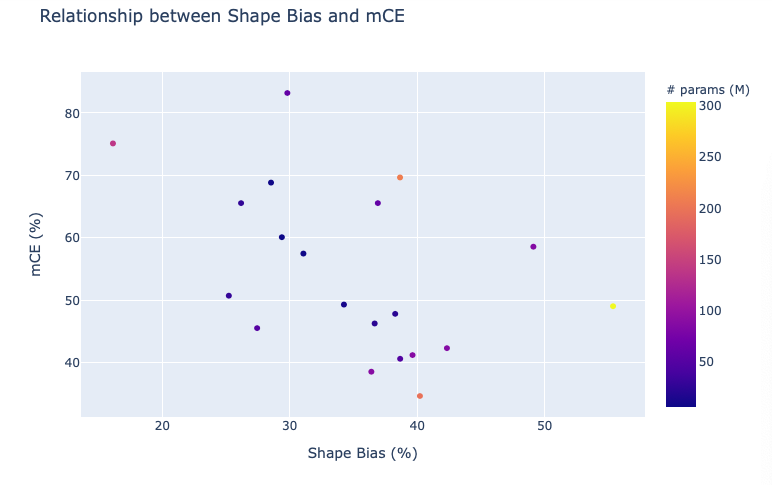}
    \caption{Relationship between mean Corruption Error and Shape Bias}
    \label{fig:graph3}
\end{figure*}

\begin{figure*}
\textbf{Note:} Please visit our \href{https://www.notion.so/Experiment-Results-30ffa0fbea9f4738ae9e0be45a0b80be}{Notion} to see our reported numbers for every subclass in ImageNet-C for severity 2 and severity 5. Camera ready paper will present these numbers in table/visualizations format in this appendix.
\end{figure*}



\end{document}